\PassOptionsToPackage{table}{xcolor}

\documentclass{article} 
\usepackage{xcolor}
\usepackage{iclr2024_conference,times}

\usepackage{color}
\usepackage{tabularray}
\usepackage{lscape} 

\usepackage{amsmath,amsfonts,bm}

\def\eqref#1{equation~\ref{#1}}

\def\1{\bm{1}}

\DeclareMathAlphabet{\mathsfit}{\encodingdefault}{\sfdefault}{m}{sl}
\SetMathAlphabet{\mathsfit}{bold}{\encodingdefault}{\sfdefault}{bx}{n}

\DeclareMathOperator*{\argmax}{arg\,max}

\usepackage{hyperref}
\usepackage{url}

\usepackage{multirow}
\usepackage{booktabs}
\usepackage{xcolor}
\usepackage{tcolorbox}
\usepackage[T1]{fontenc}
\usepackage{subcaption}
\usepackage{ifthen}
\usepackage{pgfplots}
\pgfplotsset{compat=1.17}
\usepackage{amsmath}
\usepackage{multicol}
\usepackage[font=small,labelfont=bf]{caption}
\usepackage{pifont}
\usepackage{nccmath}
\usepackage{tikz}
\usepackage[]{mdframed}
\usepackage{spverbatim}
\usepackage{fancyvrb}
\usepackage{fvextra}

\definecolor{greenolive}{RGB}{46,154,85}
\definecolor{coolgray}{RGB}{200,200,200}

\definecolor{DiseaseColor}{RGB}{93, 232, 230}       
\definecolor{DrugColor}{RGB}{255, 135, 90}          
\definecolor{ProcedureColor}{RGB}{255, 195, 111}    
\definecolor{LabColor}{RGB}{180, 235, 176}          
\definecolor{DosColor}{RGB}{158, 146, 239}          
\definecolor{FreqColor}{RGB}{188, 128, 240}         

\newcommand{\entitydis}[1]{\tikz[baseline={(a.base)}]\node[draw=none,rounded corners=0.5ex,fill=DiseaseColor!40!white,inner sep=3pt](a){#1~~\textbf{\tiny{[DIS]}}};}

\newcommand{\entitydrug}[1]{\tikz[baseline={(a.base)}]\node[draw=none,rounded corners=0.5ex,fill=DrugColor!40!white,inner sep=3pt](a){#1~~\textbf{\tiny{[DRUG]}}};}

\newcommand{\entityproc}[1]{\tikz[baseline={(a.base)}]\node[draw=none,rounded corners=0.5ex,fill=ProcedureColor!40!white,inner sep=3pt](a){#1~~\textbf{\tiny{[PROC]}}};}

\newcommand{\entitylab}[1]{\tikz[baseline={(a.base)}]\node[draw=none,rounded corners=0.5ex,fill=LabColor!40!white,inner sep=3pt](a){#1~~\textbf{\tiny{[LAB]}}};}

\title{}

\author{Wadood M Abdul,~ Marco AF Pimentel,~ Muhammad Umar Salman,~ Tathagata Raha, \\
\textbf{Clément Christophe,~ Praveen K Kanithi,~ Nasir Hayat,~ Ronnie Rajan,~ Shadab Khan} \\
M42 \\
Abu Dhabi,~ UAE\\
\texttt{\{wabdul, mpimentel, musalman, traha, cchristophe, pkanithi,} \\ \texttt{nhayat, rrajan, skhan\}@m42.ae} \\
}

\iclrfinalcopy 
\begin{document}

\begin{figure}[t]
\centering
\vspace{-.5cm}
\includegraphics[width=\linewidth,trim={2cm 5.8cm, 2cm, 5.2cm},clip]{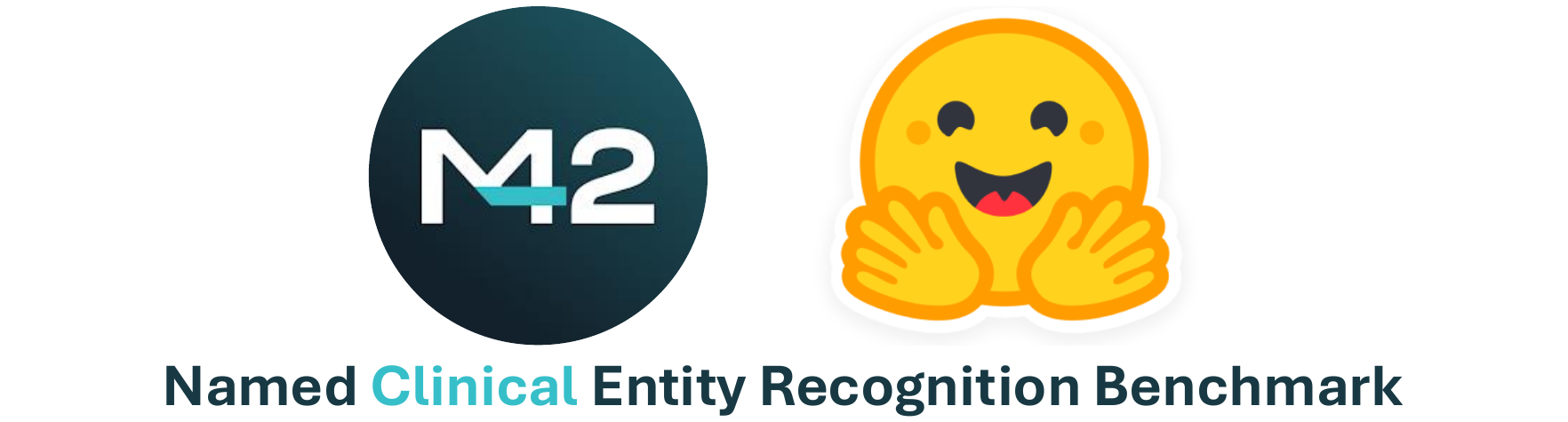}
\vspace{-1.2cm}
\label{fig:logo}
\end{figure}

\maketitle

\begin{abstract}
This technical report introduces a Named Clinical Entity Recognition Benchmark for evaluating language models in healthcare, addressing the crucial natural language processing (NLP) task of extracting structured information from clinical narratives to support applications like automated coding, clinical trial cohort identification, and clinical decision support. 

The leaderboard provides a standardized platform for assessing diverse language models, including encoder and decoder architectures, on their ability to identify and classify clinical entities across multiple medical domains. A curated collection of openly available clinical datasets is utilized, encompassing entities such as diseases, symptoms, medications, procedures, and laboratory measurements. Importantly, these entities are standardized according to the Observational Medical Outcomes Partnership (OMOP) Common Data Model, ensuring consistency and interoperability across different healthcare systems and datasets, and a comprehensive evaluation of model performance. Performance of models is primarily assessed using the F1-score, and it is complemented by various assessment modes to provide comprehensive insights into model performance. The report also includes a brief analysis of models evaluated to date, highlighting observed trends and limitations.

By establishing this benchmarking framework, the leaderboard aims to promote transparency, facilitate comparative analyses, and drive innovation in clinical entity recognition tasks, addressing the need for robust evaluation methods in healthcare NLP.  



\small{Leaderboard available at \href{https://huggingface.co/spaces/m42-health/clinical_ner_leaderboard}{\textbf{https://huggingface.co/m42-health/clinical\_ner\_leaderboard}}}.
\end{abstract}

\section{Introduction}
\label{sec:intro}
Named Entity Recognition (NER) in the clinical domain is a fundamental task in medical natural language processing (NLP), playing a crucial role in extracting structured information from unstructured clinical narratives. The ability to identify and classify entities such as diseases, symptoms, medications, and procedures within clinical texts is essential for a wide range of downstream applications \citep{Pradhan2015-bq, Stubbs2015-xj}. These applications include clinical decision support systems, where identified entities can trigger relevant alerts and/or recommendations; automated coding for billing and administrative purposes; and cohort identification for clinical trials, enabling rapid patient recruitment based on specific clinical criteria \citep{Savova2010-rb}. 

Additionally, as the volume of electronic health records (EHRs) continues to grow, efficient and accurate extraction of clinically relevant information becomes increasingly vital for both patient care and medical research \citep{Hossain2023-nl}. Accurate NER systems can significantly improve the quality of data available for clinical research, facilitate the development of precision medicine approaches, and enhance the overall efficiency of healthcare delivery \citep{Shivade2014-st, Hossain2023-nl}. 

Assessing the performance of NER tasks in the clinical domain, however, presents several challenges \citep{Kundeti2016-fb}. The inherent complexity and variability of medical terminology, coupled with the highly context-dependent nature of clinical language, make it difficult to develop universally effective NER models. Moreover, the quality of annotations in available datasets can vary significantly, affecting the reliability of performance evaluations \citep{Kundeti2016-fb, Menasalvas2016-zn, Wu2020-gw}. The scarcity of large, diverse, and well-annotated clinical datasets further complicates the assessment process, as models may perform inconsistently across different medical subdomains or institution-specific terminologies \citep{Neveol2018-md, Wu2020-gw, Niero2023-ay}.

Recent advancements in language models, particularly Large Language Models (LLMs), have shown promising results in various NLP tasks, including clinical NER \citep{Sun2021-ba, Chen2023-fp, Zhang2024-yy}. However, the lack of a standardized evaluation framework makes it challenging to compare the performance of these models objectively and consistently across different studies and datasets \citep{Peng2019-tc, Wu2020-gw, Gu2022-ai}.

To address these challenges, we present a comprehensive Named Clinical Entity Recognition (Clinical NER) Leaderboard. This leaderboard provides a standardized platform for evaluating and benchmarking the performance of various language models on clinical NER tasks. By utilizing a curated collection of openly available clinical datasets and implementing consistent evaluation metrics, our leaderboard aims to foster transparency, facilitate comparative analysis, and drive innovation in the field of clinical NER.

The key contributions of this work are summarized as follows:

\begin{itemize}
\item \textbf{Standardized evaluation framework:} We introduce a comprehensive Clinical NER Leaderboard, which provides a consistent and transparent platform for evaluating and benchmarking the performance of various language models (encoder, decoder \& gliner) on clinical NER tasks.

\item \textbf{Curated dataset collection with common standards:} The leaderboard makes use of a curated collection of openly-available clinical datasets, where entity standardization was performed using the OMOP Common Data Model standard, which ensuring that the evaluation is robust, consistent, and reflective of the diverse and context-dependent nature of clinical language.

\item \textbf{Consistent evaluation metrics:} We implement standardized evaluation metrics, allowing for objective and comparable assessments of NER models across different studies and datasets.

\item \textbf{Comparative analysis:} By providing a centralized and transparent platform, our leaderboard enables researchers to conduct comparative analyses, promoting innovation and driving progress in clinical NER research. 
\end{itemize}

These contributions, ultimately, aim to advance the field of clinical NER by addressing existing challenges and promoting the development of more accurate, reliable, and universally applicable models in healthcare applications.

\section{Related work}
\label{sec:background}

Unlike general domains, where benchmarks like GLUE \citep{Wang2018-xm} and SuperGLUE \citep{Wang2019-vb} are well-established, the biomedical field lacks equivalent resources \citep{Kanithi2024-ty}. Over the years, the field of biomedical NLP has seen the development and release of numerous datasets, often stemming from shared tasks such as BioCreative \citep{Li2016-gq}, BioNLP \citep{Demner-Fushman2024-aa}, and SemEval \citep{Ojha2024-gf}. While the focus of these datasets has evolved from simple tasks like NER to other tasks such as relation extraction and question answering, there remains a significant gap in the availability of benchmarks and leaderboards for medical and clinical NLP.

Researchers have extensively explored the use of shared language representations to capture the semantics of biomedical text, often applying these models across a range of tasks in the field \citep{Peng2019-tc}. A common approach involves transfer learning, where models are pretrained on extensive biomedical corpora and then fine-tuned for specific tasks like NER and relation extraction. BioBERT \citep{Lee2019-pb} and BioELMo \citep{Jin2019-oy} are notable examples of these approaches. These efforts have typically involved individual models evaluated in isolation, without the benefit of standardized benchmarks or leaderboards to facilitate broader comparison and validation across different approaches in medical and clinical NLP.

BLURB (Biomedical Language Understanding and Reasoning Benchmark) is one of the few benchmarks in the biomedical field, which spans multiple tasks beyond NER \citep{Gu2022-ai}. \cite{Peng2019-tc} also introduced the Biomedical Language Understanding Evaluation (BLUE) benchmark consisting of six tasks that cover both biomedical and clinical texts with different datasets. While these benchmarks provide a broad coverage of tasks, the methods and metrics used for NER tasks are not clearly detailed, and the number of domains and entities covered in the datasets is limited. Additionally, more recent approaches, such as generative models, are not included in the benchmark, indicating a gap in its ability to fully assess the latest advancements in the field \citep{Chen2023-fp}. 

While comprehensive evaluation frameworks like MEDIC \citep{Kanithi2024-ty} assess a broad range of clinical NLP tasks, in this paper we focus exclusively on NER tasks, allowing for a more detailed examination of how models are assessed and performance metrics computed. By narrowing our scope to NER, we can delve deeper into the intricacies of model evaluation, ensuring that the metrics used provide a comprehensive understanding of a model's capability to accurately identify and classify entities within the medical and clinical domains. This focused approach also enables us to explore the latest trends in utilizing large language models for diverse NER tasks, providing a platform to compare the performance of different model architectures. Finally, this work also emphasizes the importance of standardizing entities across models and datasets according to widely accepted standards, a critical aspect that has been insufficiently addressed in previous works. Overall, we aim to highlight the strengths and limitations of various models, offering insights into how these models perform in specialized tasks that are crucial for advancing biomedical NLP.  

\section{The Clinical NER Benchmark}
\label{sec:cner}
To address the challenges in evaluating clinical NER models, we have developed a benchmark that provides a standardized platform for assessing performance. This benchmark consists of the following key components: it contains a common evaluation methodology that employs well-established evaluation metrics, primarily focusing on the F1-score; it employs terminology standardization of the clinical entities included in our evaluation, which ensures consistency and interoperability; and it includes a curated collection of openly available medical benchmark datasets, encompassing a broad spectrum of medical entities. In the subsections below, we first elucidate the problem and then elaborate on the components in the following subsections.

\subsection{Named-Entity Recognition Task}

NER is a crucial task in biomedical NLP that aims to identify and classify medical entities in unstructured clinical text. Mathematically, we can formulate the NER task as follows. Given an input sequence of tokens $X = (x_1, x_2, \ldots, x_n)$, where each $x_i$ represents a token (a word or sub-word) in clinical text, the goal is to assign a corresponding sequence of labels $Y = (y_1, y_2, \ldots, y_n)$, where each $y_i$ belongs to a predefined set of clinical entity types $E \cup \{O\}$, with $O$ representing the ``Outside'' label for tokens that are not part of any medical entity.

Formally, we can express this as a function $f : X \rightarrow Y$, where $X$ is the space of all possible input sequences of text, and $Y$ is the space of all possible clinical label sequences.



The set of clinical entity types $E$ typically includes categories such as $E=\{\text{DIS},\text{PROC},\text{DRUG},\ldots\}$,  
where, for example: 
\begin{itemize}
    \item \emph{DIS} corresponds to medical conditions or disorders,
    \item \emph{PROC} includes medical procedures or interventions,
    \item \emph{DRUG} relates to medications.
\end{itemize} 

The NER task can be viewed as a sequence labeling problem, where we aim to maximize the conditional probability $P(Y|X)$, i.e., $\argmax_Y P(Y|X)$.


This probability can be modeled using various approaches, such as Conditional Random Fields (CRFs), or neural network architectures like Bidirectional Long Short-Term Memory (BiLSTM) networks or Transformer-based models fine-tuned on clinical corpora \citep{Wu2020-gw}.

To illustrate the clinical NER task, consider the following example:

\begin{mdframed}[backgroundcolor=black!4,rightline=true,leftline=true]
Patient presents with \entitydis{acute myocardial infarction} and is prescribed \entitydrug{aspirin} until \entityproc{angioplasty} is performed.
\end{mdframed}

The input sequence $X$ is:

\verb|Patient presents with acute myocardial infarction and is|
\verb|prescribed aspirin until angioplasty is performed|

The corresponding label sequence $Y$ (assuming each word is a token):

\verb|O O O B-DIS I-DIS I-DIS O O|
\verb|O B-DRUG O B-PROC O O|

Where \emph{B-*} indicates the beginning of an entity, \emph{I-*} indicates the continuation (inside) of an entity, and \emph{O} indicates tokens outside of clinical entities of interest.

This example demonstrates how the clinical NER task assigns labels to each token in the input sequence, identifying ``acute myocardial infarction'' as a disease, ``aspirin'' as a drug, and ``angioplasty'' as a procedure.

\subsection{Evaluation metrics}
\label{sec:metrics}

The performance of clinical NER models, which aim to optimize $P(Y|X)$ as shown in equation (3), is evaluated using two types of metrics: token-based and span-based. Both types utilize precision, recall, and F1-score, but they differ in how they define true positives (TP), false positives (FP), and false negatives (FN).

\subsubsection{Token-based Metrics}

Token-based metrics evaluate the model's performance at the individual token level. For each token $x_i$ in the input sequence $X$, we compare the predicted label $\hat{y}_i$ with the true label $y_i$. Let $TP_t$, $FP_t$, and $FN_t$ represent token-level true positives, false positives, and false negatives, respectively. Then:

\begin{fleqn}
\begin{align}
&&&&&& \text{Precision}_t &= \frac{TP_t}{TP_t + FP_t} &&&&&&&&&&&&&&&&&&&&&&&&&&&& \\
&&&&&& \text{Recall}_t    &= \frac{TP_t}{TP_t + FN_t} &&&&&&&&&&&&&&&&&&&&&&&&&&&& \\
&&&&&& \text{F1-score}_t  &= 2 \cdot \frac{\text{Precision}_t \cdot \text{Recall}_t}{\text{Precision}_t + \text{Recall}_t} &&&&&&&&&&&&&&&&&&&&&&&&&&&&
\end{align}
\end{fleqn}

The above metrics can be calculated either globally or on a per entity type basis, thus giving us two possible metrics:
\begin{itemize}
    \item Micro Average: The $TP_t$, $FP_t$, and $FN_t$ values are calculated globally to get the final precision, recall and F1 values.
    \item Macro Average: The precision, recall and F1 are calculated for each entity type and then averaged without any weightage.
\end{itemize}

With this token-based approach, we have a broad idea of the performance of the model at the token level. However, it may misrepresent the performance at the entity level when the entity includes more than 1 token (which may be more relevant for certain applications). In addition, depending on the annotations of certain datasets, we may not want to penalize a model for a "partial" match with a certain entity.

\subsubsection{Span-based Metrics}

Span-based metrics evaluate the model's performance at the entity level, considering full or partial matches. These metrics are particularly important in clinical NER, as they reflect the model's ability to identify complete medical entities. Let $TP_s$, $FP_s$, and $FN_s$ represent span-level true positives, false positives, and false negatives, respectively. We define:

\begin{itemize}
    \item Exact Match: The predicted entity spans exactly match the true entity span's boundary and label.
    \item Partial Match: The predicted entity spans overlap with the true entity span's boundary and exactly matches the label.
\end{itemize}

Based on the criteria above, each predicted or true span can be classified as $Correct$, $Incorrect$, $Missed$, $Spurious$ (see Table~\ref{tab:span-metrics}).

\begin{table}[t]
\caption{\textbf{Exact and partial span metric calculations.} Each predicted span can be attributed to each class depending on exact or partial matches.}
\label{tab:span-metrics}
\small
\begin{tabular}{p{0.15\linewidth}p{0.38\linewidth}p{0.38\linewidth}}
\toprule
\textbf{Span Class} & \textbf{Exact} & \textbf{Partial} \\
\midrule
Correct &
  The predicted and true span's boundary and label match exactly &
  The predicted and true span's label matches exactly and the boundary has some overlap \\ 
\rowcolor{coolgray!20} Incorrect &
  There is a mismatch in either the boundary or label between the predicted and true span &
  There is an overlap in the boundary of predicted and true span but a mismatch in the label \\ 
Missed              & For a given True span, there is no predicted span that has overlap with it & For a given True span, there is no predicted span that has overlap with it \\
\rowcolor{coolgray!20} Spurious            & For a given predicted span, there is no true span that has an exact overlap with it & For a given predicted span, there is no true span that has any overlap with it \\ 
\bottomrule
\end{tabular}
\end{table}

Using the above classifications, we have 

\begin{fleqn}
\begin{align}
&&&&&& \text{}FP_s &= {Incorrect + Spurious} &&&&&&&&&&&&&&&&&&&&&&&&&&&& \\
&&&&&& \text{}FN_s &= {Incorrect + Missed} &&&&&&&&&&&&&&&&&&&&&&&&&&&&
\end{align}
\end{fleqn}

Then, we calculate:

\begin{fleqn}
\begin{align}
&&&&&& \text{Precision}_s &= \frac{TP_s}{TP_s + FP_s} &&&&&&&&&&&&&&&&&&&&&&&&&&&& \\
&&&&&& \text{Recall}_s    &= \frac{TP_s}{TP_s + FN_s} &&&&&&&&&&&&&&&&&&&&&&&&&&&& \\
&&&&&& \text{F1-score}_s  &= 2 \cdot \frac{\text{Precision}_s \cdot \text{Recall}_s}{\text{Precision}_s + \text{Recall}_s} &&&&&&&&&&&&&&&&&&&&&&&&&&&&
\end{align}
\end{fleqn}

Strict span based evaluation may be more applicable in applications like de-identifying PII, where as partial span based evaluation is desirable when we have leading/following words that do not change the entity's meaning. 

\subsubsection{Working Example}

Consider the following example, with the following entities (i.e., true labels):

\begin{mdframed}[backgroundcolor=black!4,rightline=true,leftline=true]
The patient's \entityproc{chest X-ray} showed \entitydis{pneumonia}, and \entitylab{blood cultures} were ordered to rule out \entitydis{sepsis}. Patient has no \entitydis{diabetes}. \entitydrug{Levofloxacin} was prescribed for treatment."
\end{mdframed}

Assume the predicted labels are as follows:

\begin{mdframed}[backgroundcolor=black!4,rightline=true,leftline=true]
The patient's chest \entityproc{X-ray} showed \entitydis{pneumonia}, and \entitylab{blood cultures} were ordered to rule out sepsis. Patient has no \entitydis{diabetes}. \entitydrug{Levofloxacin} was prescribed for \entityproc{treatment}."
\end{mdframed}

Token-based evaluation (Micro Average):
\begin{itemize}
    \item TP$_t$ = 6 (X-ray, pneumonia, blood, cultures, diabetes, Levofloxacin)
    \item FP$_t$ = 1 (treatment)
    \item FN$_t$ = 2 (chest, sepsis)
    \item F1-score$_t$ = \textbf{0.80}
\end{itemize}

Token-based evaluation (Macro Average):
\begin{itemize}
    \item TP$_t$ = {PROC: 1, DIS: 2, DRUG: 1, LAB: 2 }
    \item FP$_t$ = {PROC: 1, DIS: 0, DRUG: 0, LAB: 0 }
    \item FN$_t$ = {PROC: 1, DIS: 1, DRUG: 0, LAB: 0 }
    \item Precision$_t$ = {PROC: 0.5, DIS: 1, DRUG: 1, LAB: 1 }
    \item Recall$_t$ = {PROC: 0.5, DIS: 0.66, DRUG: 1, LAB: 1 }
    \item F1$_t$ = {PROC: 0.5, DIS: 0.8, DRUG: 1, LAB: 1 }
    \item Final F1-score$_t$ = \textbf{0.82}
\end{itemize}

Span-based evaluation (Exact Match):
\begin{itemize}
    \item TP$_s$ = 4 (pneumonia, blood cultures, diabetes, Levofloxacin)
    \item FP$_s$ = 2 (chest X-ray, treatment)
    \item FN$_s$ = 2 (chest X-ray, sepsis)
    \item F1-score$_s$ = \textbf{0.66}
\end{itemize}

Span-based evaluation (Partial Match):
\begin{itemize}
    \item TP$_s$ = 5 (chest X-ray, pneumonia, blood cultures, diabetes, Levofloxacin)
    \item FP$_s$ = 1 (treatment)
    \item FN$_s$ = 1 (sepsis)
    \item F1-score$_s$ = \textbf{0.83}
\end{itemize}

This example demonstrates how token-based and span-based metrics can provide different perspectives on model performance. Span-based metrics, in particular, reveal issues with entity boundary detection, particularly for the procedure entity. The partial match evaluation shows better performance than the exact match, indicating that the model is generally identifying the correct entities but sometimes struggles with precise boundaries.

For our evaluation framework we consider the \emph{Macro Average} token-based metrics and the \emph{Partial Match} for our span-based metrics.

The variety of entity types demonstrated in this example (\emph{procedure}, \emph{disease}, \emph{lab test}, \emph{drug}) highlights the complexity of clinical NER tasks. To ensure consistency across different NER systems and to facilitate interoperability in clinical applications, it is crucial to establish a standardized terminology for entity types. This standardization not only aids in the accurate evaluation of NER models but also enhances the utility of extracted information in downstream tasks such as clinical decision support systems. The following section delves into the importance and implementation of common terminologies in clinical NER.

\subsection{Common terminology}
\label{sec:terminology}
Standardization of medical terminology is a critical requirement for the effective development and deployment of clinical NLP systems. In the medical field, the proliferation of institution-specific vocabularies, coding systems, and ontologies has long posed a significant challenge for data integration, interoperability, and the generalization of NLP models across different healthcare settings \citep{Iroju2015-yr}.

To address this issue, the Observational Medical Outcomes Partnership (OMOP) Common Data Model (CDM) has emerged as a widely adopted standard for harmonizing clinical data \citep{Observational-Health-Data-Sciences2021-ej}. The OMOP CDM provides a standardized framework for organizing and representing a wide range of medical concepts, including diagnoses, procedures, medications, laboratory tests, and demographic information. By mapping diverse source terminologies to the common OMOP concepts and vocabularies, the model enables seamless integration and analysis of data from multiple institutions and data sources.

The importance of terminology standardization is particularly evident in the context of clinical NER, where the accurate identification and classification of medical entities are crucial for downstream applications such as clinical decision support, automated coding, and cohort identification. Inconsistent or ambiguous representations of these entities can lead to significant errors and performance degradation in NER models \citep{Kundeti2016-fb, Klug2024-gy}.

In the development of our Clinical NER Benchmark, we have leveraged the OMOP Common Data Model to standardize the medical entities included in the evaluation datasets. By aligning the entities to the OMOP standard vocabularies, we ensure that the benchmark provides a consistent and interoperable representation of clinical concepts, facilitating fair comparisons of NER model performance across diverse datasets and healthcare settings. Furthermore, we propose two additional domains - genes and gene variants - to cover genomic data, aligning with the OMOP CDM extension for storing genetic information, thus enhancing the benchmark's applicability to precision medicine and genomics research \citep{Shin2019-xw}. Table~\ref{tab:omop-domains} provides an overview of these domains, including brief descriptions and examples for each entity type.

\begin{table}[t]
\caption{\textbf{Standard clinical entities.} Brief description of the OMOP domains used in the Clinical NER Benchmark.}
\label{tab:omop-domains}
\small
\begin{tabular}{p{0.15\linewidth}p{0.45\linewidth}p{0.30\linewidth}}
\toprule
\textbf{Entity Type} & \textbf{Description} & \textbf{Examples} \\
\midrule
Conditions & Medical diagnoses, symptoms, or clinical findings & Pneumonia, Hypertension, Chest pain \\
\rowcolor{coolgray!20} Procedures & Medical, surgical, or diagnostic interventions & Appendectomy, MRI scan, Blood transfusion \\
Drugs & Medications, therapeutic agents, or substances used for treatment & Aspirin, Insulin, Amoxicillin \\
\rowcolor{coolgray!20} Measurements & Laboratory tests, vital signs, or other quantifiable clinical observations & Blood glucose level, Body temperature, Serum creatinine \\
Genes & Specific genes or genetic loci relevant to clinical contexts & BRCA1, TP53, EGFR \\
\rowcolor{coolgray!20} Gene Variants & Specific alterations or mutations in genes & BRAF V600E, EGFR T790M, KRAS G12D \\
\bottomrule
\end{tabular}
\end{table}

By incorporating these OMOP domains, our Clinical NER Benchmark provides a comprehensive framework for evaluating NER models across a diverse range of clinical entities. This approach not only ensures broad coverage of medically relevant concepts but also facilitates the benchmark's applicability to various clinical specialties and research areas, including oncology, pharmacogenomics, and rare genetic disorders. Importantly, the use of the OMOP CDM as our standardization framework ensures the scalability and future-proofing of our benchmark. Additional entity types or domains can be seamlessly integrated into the benchmark in the future, following a careful mapping process to align with OMOP standards. This extensibility allows our benchmark to evolve alongside advancements in medical knowledge and changing clinical information needs, maintaining its relevance and comprehensiveness over time.


\subsection{Datasets}
\label{sec:datasets}

Four publicly-available datasets have been included in our benchmark. They are summarized in Table~\ref{tab:datasets}.

\paragraph{NCBI} The NCBI Disease corpus includes mention and concept level annotations on 100 PubMed abstracts \citep{Dogan2014NCBIDC}. It covers annotations of diseases.

\paragraph{CHIA} This is large, annotated corpus of patient eligibility criteria extracted from 194 registered clinical trials \citep{kury2020chia}. Annotations cover 15 entity types (according to OMOP domains), including conditions, drugs, procedures, and measurements.

\paragraph{BC5CDR} The BC5CDR corpus contains PubMed articles with human annotations of all chemicals and diseases \citep{DBLP:journals/biodb/LiSJSWLDMWL16}. 

\paragraph{BIORED} The BIORED corpus includes a set of PubMed abstracts with annotations of multiple entity types, including genes/proteins, diseases, and chemicals \citep{DBLP:journals/corr/abs-2204-04263}.

\begin{table}[t]
\caption{\textbf{Summary of publicly available datasets.} The standard entities that are included in each dataset is also shown here. For detailed entity type mapping refer \ref{tab:Entity Standardization Mapping}}
\label{tab:datasets}
\small
\begin{tabular}{lccll}
\toprule
\textbf{Dataset} & \textbf{\# samples} & \textbf{\# annotations} & \textbf{Entity types} & \textbf{Corpus} \\
\midrule
NCBI & 100 & 960 & Condition & PubMed \\
\rowcolor{coolgray!20} CHIA & 194 & 3,981 & Condition, Procedure, Measurement, Drug & Clinical Trials \\
BC5CDR & 500 & 9,928 & Condition, Drug & PubMed \\
\rowcolor{coolgray!20} BIORED & 100 & 3,535 & Condition, Drug, Gene, Gene variant & PubMed \\
\bottomrule
\end{tabular}
\end{table}

The above datasets were adapted to align with our evaluation framework by mapping the annotations to clinically relevant entity types, as defined by the OMOP CDM. Entity types not included in the framework were omitted due to the limited availability of datasets with sufficient annotations for those entities. To ensure consistency, the retained clinical entity types were standardized across all datasets, resulting in a final set of six clinical entity types, as detailed in Table~\ref{tab:omop-domains}.

\section{Results and Analysis}
\label{sec:results}

We performed an analysis of the performance of various models evaluated on the proposed benchmarks and included on our leaderboard, showcasing the outcomes of the models assessed to date, with additional models planned to be incorporated in future iterations.

\subsection{Model Diversity}
The analysis encompassed a diverse range of model architectures, including encoder-only, decoder-only, and the recently proposed GLiNER models \citep{zaratiana2023gliner}. These models varied in size, pre-training data, and whether they underwent fine-tuning for the NER task. Table \ref{tab:models} provides a summary of the models evaluated in this study, highlighting their architectural differences and key characteristics.

\begin{table}[t]
\caption{\textbf{Current Models on the Leaderboard.} Models varying in architecture, training data scope and sizes are currently included on the leaderboard.}
\label{tab:models}
\small
\centering
\begin{tabular}{lllr}
\toprule
Model & Architecture & Type & \#Params (M) \\
\midrule
Universal-NER/UniNER-7B-type-sup & Decoder & fine-tuned & 7000 \\
\rowcolor{coolgray!20} Universal-NER/UniNER-7B-all & Decoder & fine-tuned & 7000 \\
knowledgator/gliner-multitask-large-v0.5 & GLiNER Encoder & zero-shot & 304 \\
\rowcolor{coolgray!20} gliner-community/gliner\_large-v2.5 & GLiNER Encoder & zero-shot & 304 \\
urchade/gliner\_large\_bio-v0.1 & GLiNER Encoder & zero-shot & 304 \\
\rowcolor{coolgray!20} Universal-NER/UniNER-7B-type & Decoder & zero-shot & 7000 \\
openai/gpt-4o-2024-05-13 & Decoder & zero-shot & - \\
\rowcolor{coolgray!20} EmergentMethods/gliner\_large\_news-v2.1 & GLiNER Encoder & zero-shot & 304 \\
urchade/gliner\_large-v2.1 & GLiNER Encoder & zero-shot & 304 \\
\rowcolor{coolgray!20} openai/gpt-4o-mini-2024-07-18 & Decoder & zero-shot & - \\
numind/NuNER\_Zero & GLiNER Encoder & zero-shot & 304 \\
\rowcolor{coolgray!20} numind/NuNER\_Zero-span & GLiNER Encoder & zero-shot & 304 \\
meta-llama/Meta-Llama-3.1-8B-Instruct & Decoder & zero-shot & 8030 \\
\rowcolor{coolgray!20} meta-llama/Meta-Llama-3-8B-Instruct & Decoder & zero-shot & 8030 \\
meta-llama/Meta-Llama-3-70B-Instruct & Decoder & zero-shot & 70000 \\
\rowcolor{coolgray!20} alvaroalon2/biobert\_diseases\_ner & Encoder & fine-tuned & 110 \\
bioformers/bioformer-8L-ncbi-disease & Encoder & fine-tuned & 43 \\
\rowcolor{coolgray!20} mistralai/Mixtral-8x7B-Instruct-v0.1 & Decoder & zero-shot & 45000 \\
\bottomrule
\end{tabular}
\end{table}

The different model architectures included in the leaderboard are:
\begin{itemize}
    \item \textbf{Encoder}: The standard token classification model built on top of transformer encoder architecture.
    \item \textbf{Decoder}: Autoregressive token generation models based on the transformer decoder architecture.
    \item \textbf{GLiNER Encoder}: An enhancement on the transformer encoder architecture that uses similarity between span and entity embeddings.
\end{itemize}

The models also vary in the scope of training data used. The models that have been exposed to any of the training data on the benchmark have been categorised as Type: 'fine-tuned' and the models with no exposure to the training data from the benchmark have been categorised as Type:'zero-shot'\footnote{Note: Some of the zero-shot models may have exposure to the benchmark's clinical entities by being trained on open source or synthetically generated datasets that have similar entities.} .

The inclusion of this diverse set of models allows for a comprehensive evaluation of different approaches to clinical NER, spanning from general-purpose language models (e.g., LLMs) to those specifically designed for token classification tasks.

\subsection{Entity-specific Performance}

\begin{figure}[t]
    \centering
    \begin{subfigure}{0.495\textwidth}
        \includegraphics[width=\linewidth]{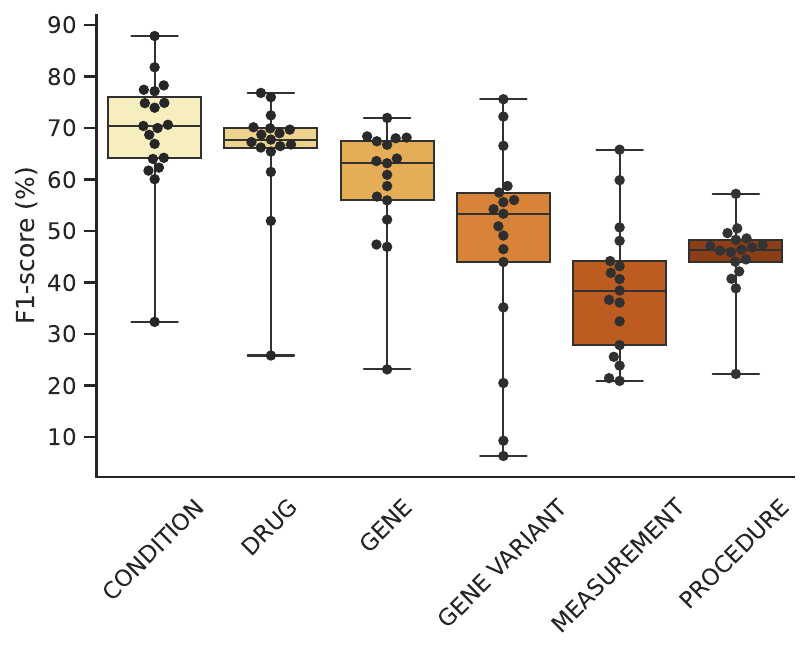}
        \caption{Token-based}
    \end{subfigure}
    \begin{subfigure}{0.495\textwidth}
        \includegraphics[width=\linewidth]{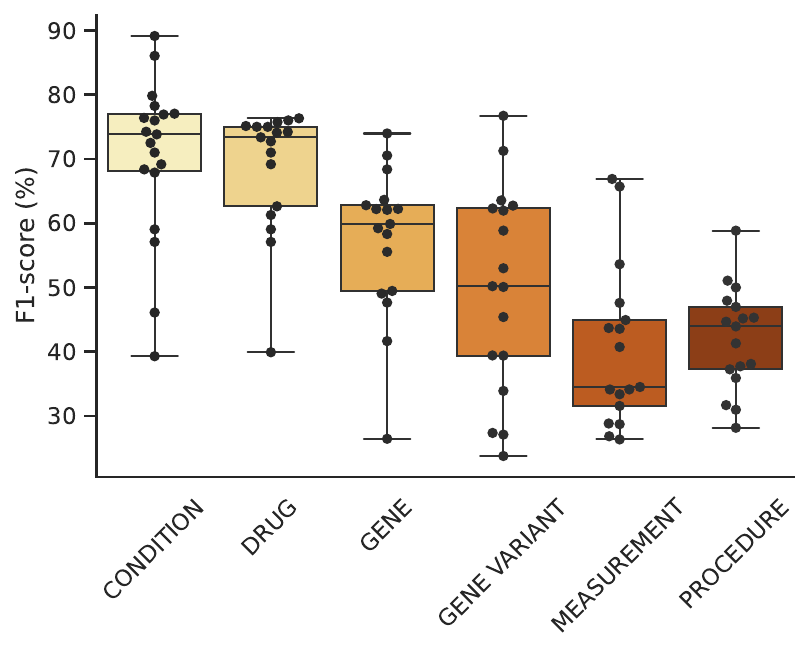}
        \caption{Span-based}
    \end{subfigure}
    \caption{\textbf{Overall performance of models across six clinical entities}. Box plots represent F1-scores of various models across the clinical entity types for each metric approach: token-based (left), and span-based. Each dots represent the performance of a model. }
    \label{fig:perf-per-entity}
\end{figure}

Figure~\ref{fig:perf-per-entity} shows the overall performance of all models for each entity type using both span-based and token-based metrics. 

A notable observation from this analysis is the higher performance (F1-score) for condition and drug entities compared to other entity types, which is observed for both span-based and token-based approaches. This trend may be attributed to the prevalence and consistency of these entity types in clinical texts, as well as their potentially more standardized representation in medical terminology. This is also reflected in figure \ref{fig:span_counts} that shows the span counts for each entity type present on the leaderboard.

Interestingly, when examining the performance for a single entity type (condition) across different datasets (Figure \ref{fig:perf-condition-datasets}), we observe relatively consistent performance. This suggests that the models' ability to recognize Condition entity type (for example) may be generalizable across various clinical contexts and data sources.

\subsection{Impact of Model Size and Architecture}

\begin{figure}[t]
    \centering
    \begin{subfigure}{0.495\textwidth}
        \includegraphics[width=\linewidth]{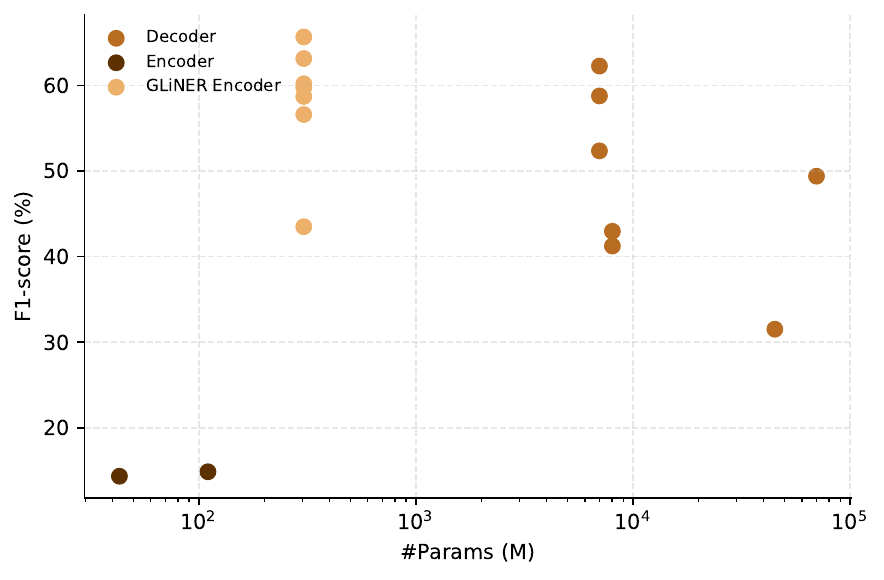}
        \caption{Size Comparision}
    \end{subfigure}
    \begin{subfigure}{0.495\textwidth}
        \includegraphics[width=\linewidth]{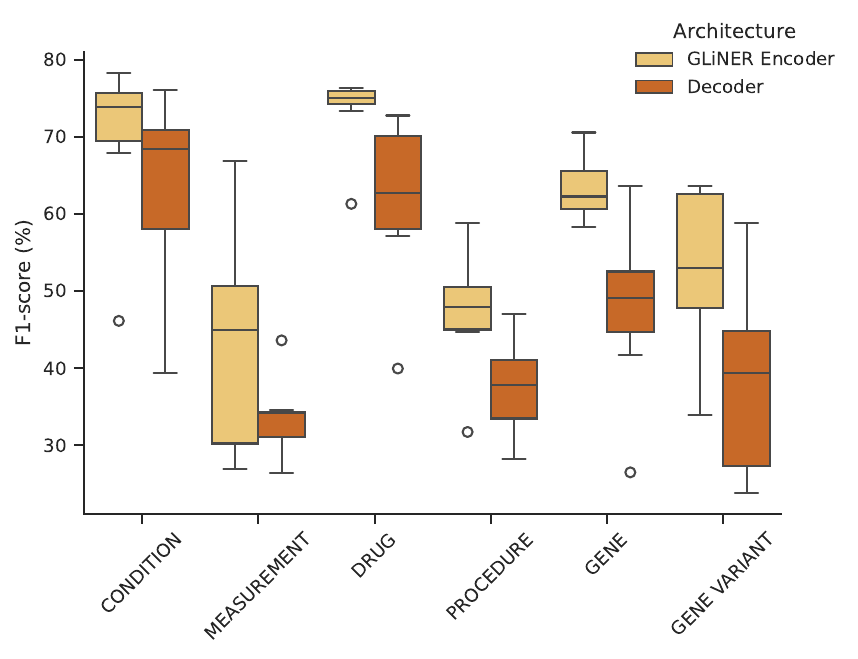}
        \caption{Architecture Comparision}
    \end{subfigure}
    \caption{\textbf{Performance across model sizes and architectures}. Both the plots represent the span based F1 scores. For size comparision (left) the average F1 score across clinical entities is used. For architecture comparision (right) only decoder and GLiNER encoder models are used. Additionally, closed source models are filtered out.   }
    \label{fig:perf-architecture}
\end{figure}

Figure~\ref{fig:perf-architecture} illustrates the performance of models according to their size and architecture. 

A key finding from this analysis is that LLMs models (i.e., decoder-only architectures) generally do not perform as well as the specialized encoder-based GLiNER architecture for the clinical NER task. This disparity in performance may be attributed to the inherent strengths of encoder-based architectures in token classification tasks, which align closely with the requirements of NER. GLiNER was designed specifically for token classification tasks, utilizing span and label embedding's similarity, this likely contributes to its strong performance in this task. Decoder models on the other hand generate tokens in an auto-regressive manner, this limits it's ability to extract accurate span information, a task which is extractive in nature.

\subsubsection{Impact of Finetuning}

\begin{figure}[!b]
    \centering
        \includegraphics[width=0.7\linewidth]{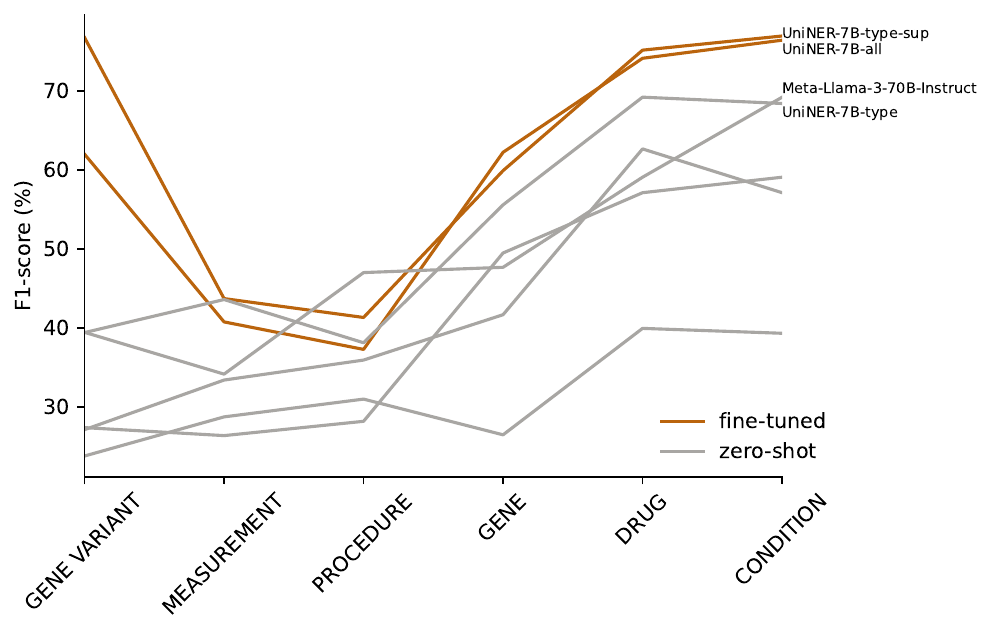}
        \caption{\textbf{Effect of Training.} Span based metrics of the open-source decoder models from the leaderboard are used here. }
    \label{fig:training-impact}
\end{figure}

Figure \ref{fig:training-impact} depicts the performance across clinical entities of fine-tuned and zero-shot models . Only the decoder architecture subset is used for this comparison as architectures like GLiNER do not have a supervised variant at the time of writing the paper.

We note that the best performance is obtained by supervised models, which is an expected result. Among the zero-shot models, in lead are Meta-Llama-3-70B-Instruct which is much larger in size and UniNER-7B-type which has been trained on task specific synthetically generated data.

\subsection{Token-based vs. Span-based Evaluation}

\begin{figure}[t]
    \centering
    \includegraphics[width=.6\linewidth]{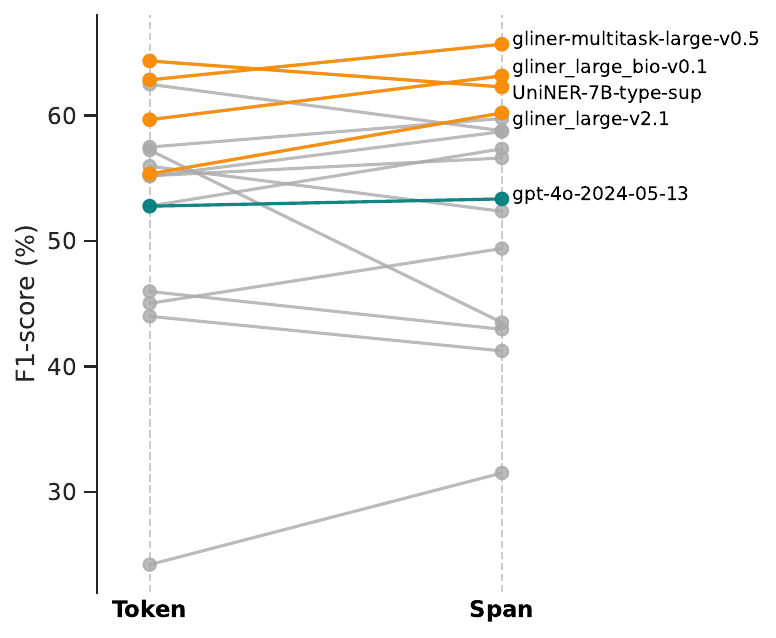}
    \caption{\textbf{Model rankings according to token-based and span-based F1-scores.} The overall (average) token (left) and span-based metrics for each model are shown. The top-4 performing models, according to the span-based F1-score, are highlighted in orange, and the performance of GPT-4o is shown in teal. Models with overall performances below 20\% are not shown.}
    \label{fig:perf-token-v-span}
\end{figure}

We have also compared token-based and span-based performance metrics for the evaluated models. While the core messages and trends derived from both evaluation approaches remain consistent, we observed differences in the absolute performance values and relative rankings of models between the two metrics (as shown in Figure~\ref{fig:perf-token-v-span}).

Token-based and span-based F1-scores reveal clear ranking distinctions between models. The figure compares the overall (average) token-based and span-based F1-scores for each model, highlighting the ranking of models according to each metric and providing insight into model performance across different evaluation approaches.

These differences highlight the importance of considering both evaluation methodologies in clinical NER tasks. Token-based metrics provide insights into the models' ability to correctly classify individual tokens, while span-based metrics offer a more holistic view of entity recognition. The disparity between these metrics underscores the complexity of clinical NER and the need for comprehensive evaluation approaches to fully understand model performance.

\section{Discussion and Conclusions}
\label{sec:discussion}

In this work, we introduce a Clinical NER Benchmark, providing a standardized framework for evaluating language models for NER tasks. Our work addresses some critical challenges in clinical NLP and offers valuable insights into model performance across various clinical domains.

A key strength of this work lies in its comprehensive approach to addressing persistent challenges in clinical NLP. First, our leaderboard tackles the issue of non-standardized medical data formats through terminology standardization. By leveraging the OMOP CDM for entity standardization, we promote consistency and interoperability across diverse healthcare systems and datasets. This standardization not only facilitates more meaningful comparisons between models but also enhances the potential for collaborative research and development in clinical NLP. Second, we have processed a set of benchmark datasets that cover various entity types and clinical domains. This diverse collection ensures a robust evaluation of model performance across different aspects of clinical narratives, providing a more comprehensive assessment of a model's capabilities in real-world healthcare scenarios. Third, our methodology for evaluation includes different criteria for computing standard metrics such as precision, recall, and F1-score, this allows for a direct comparisons with existing literature while offering comprehensive insights into model performance, addressing the multifaceted nature of entity recognition tasks.

Our evaluation of various models included on the leaderboard (to date) has yielded some important insights. GLiNER-based models have demonstrated superior performance across multiple datasets and entity types. In contrast, decoder-only architectures, used by LLMs such as Llama-3 and GPT-4o, have shown comparatively lower performance. A similar trend has been observed in other studies \citep{Chen2023-fp, Soroush2024-kq}. Furthermore, our analysis revealed that the choice of evaluation strategy—token-based or span-based—can significantly impact the ranking of models, highlighting the importance of comprehensive assessment approaches in clinical NER tasks.

With the establishment of this leaderboard, we aim to drive significant advancements in clinical NLP, with a particular focus on NER. By providing a standardized platform for evaluating diverse language models, including LLMs, we enable researchers and practitioners to benchmark their approaches against state-of-the-art performance. This transparency and comparability are crucial for driving innovation and improving the accuracy of clinical entity recognition tasks, which have far-reaching implications for applications such as clinical decision support, automated coding, and cohort identification for clinical trials.

Although our current evaluation metrics focus on traditional measures such as precision, recall, and F1-score, we recognize the potential for more refined assessment approaches. For instance, \cite{Fu2020-br} proposed an alternative methodology that defines explainable attributes of data (e.g., entity density, label consistency, token frequency) and evaluates models on distinct buckets based on these attributes. This granular approach allows for a more detailed understanding of model performance, identifying specific areas of strength and weakness. Incorporating such methodologies into future iterations of our leaderboard could provide even more actionable insights for researchers and developers, guiding targeted improvements in model architectures and training strategies.

It is important to acknowledge a significant limitation in the field of clinical NER, which is also reflected in our leaderboard: the issue of label imbalance. Clinical datasets, such as those used in this work, often exhibit a skewed distribution of entity types, with some categories being far more prevalent than others. This imbalance can lead to reporting biased model performances, where accuracy on common entities (such as conditions) may overshadow poor performance on less prevalent (annotated) clinical entities. Future work on this leaderboard and in the broader field of clinical NER should address this limitation through the development of more balanced benchmark datasets. 

We are actively working to expand the scope and utility of the  Clinical NER Leaderboard\footnote{https://huggingface.co/spaces/m42-health/clinical\_ner\_leaderboard}. Additional internal datasets are in the process of being included, which will further enhance the robustness and generalizability of model evaluations. Moreover, we enthusiastically welcome contributions from the broader research community. Whether in the form of new datasets, innovative model architectures, or improvements to the \href{https://github.com/WadoodAbdul/clinical_ner_benchmark}{clinical NER Benchmark codebase}\footnote{https://github.com/WadoodAbdul/clinical\_ner\_benchmark},  external contributions will play a crucial role in the continued evolution and relevance of this resource. To facilitate engagement, we have implemented an automatic submission form, streamlining the process for researchers to add their models to the leaderboard.

In conclusion, by addressing key challenges in data standardization and providing a platform for transparent comparison, we aim to accelerate progress in this critical domain of healthcare informatics. As we continue to refine and expand this resource, we look forward to the insights and innovations it will foster within the research community, ultimately contributing to more accurate and efficient processing of clinical narratives.

\section*{Acknowledgements}

This work was supported by M42. 


\bibliography{iclr2024_conference}

\begin{thebibliography}{34}
\providecommand{\natexlab}[1]{#1}
\providecommand{\url}[1]{\texttt{#1}}
\expandafter\ifx\csname urlstyle\endcsname\relax
  \providecommand{\doi}[1]{doi: #1}\else
  \providecommand{\doi}{doi: \begingroup \urlstyle{rm}\Url}\fi

\bibitem[Chen et~al.(2023)Chen, Sun, Liu, Jiang, Ran, Jin, Xiao, Lin, Chen, and Niu]{Chen2023-fp}
Qijie Chen, Haotong Sun, Haoyang Liu, Yinghui Jiang, Ting Ran, Xurui Jin, Xianglu Xiao, Zhimin Lin, Hongming Chen, and Zhangmin Niu.
\newblock An extensive benchmark study on biomedical text generation and mining with {ChatGPT}.
\newblock \emph{Bioinformatics}, 39\penalty0 (9), September 2023.

\bibitem[Demner-Fushman et~al.(2024)Demner-Fushman, Ananiadou, Miwa, Roberts, and Tsujii]{Demner-Fushman2024-aa}
Dina Demner-Fushman, Sophia Ananiadou, Makoto Miwa, Kirk Roberts, and Junichi Tsujii (eds.).
\newblock \emph{Proceedings of the {23rd} Workshop on Biomedical Natural Language Processing}, Bangkok, Thailand, August 2024. Association for Computational Linguistics.

\bibitem[Dogan et~al.(2014)Dogan, Leaman, and Lu]{Dogan2014NCBIDC}
Rezarta~Islamaj Dogan, Robert Leaman, and Zhiyong Lu.
\newblock Ncbi disease corpus: A resource for disease name recognition and concept normalization.
\newblock \emph{Journal of biomedical informatics}, 47:\penalty0 1--10, 2014.

\bibitem[Fu et~al.(2020)Fu, Liu, and Neubig]{Fu2020-br}
Jinlan Fu, Pengfei Liu, and Graham Neubig.
\newblock Interpretable multi-dataset evaluation for named entity recognition.
\newblock In \emph{Proceedings of the 2020 Conference on Empirical Methods in Natural Language Processing ({EMNLP})}, pp.\  6058--6069, Stroudsburg, PA, USA, November 2020. Association for Computational Linguistics.

\bibitem[Gu et~al.(2022)Gu, Tinn, Cheng, Lucas, Usuyama, Liu, Naumann, Gao, and Poon]{Gu2022-ai}
Yu~Gu, Robert Tinn, Hao Cheng, Michael Lucas, Naoto Usuyama, Xiaodong Liu, Tristan Naumann, Jianfeng Gao, and Hoifung Poon.
\newblock Domain-specific language model pretraining for biomedical natural language processing.
\newblock \emph{ACM Trans. Comput. Healthc.}, 3\penalty0 (1):\penalty0 1--23, January 2022.

\bibitem[Hossain et~al.(2023)Hossain, Rana, Higgins, Soar, Barua, Pisani, and Turner]{Hossain2023-nl}
Elias Hossain, Rajib Rana, Niall Higgins, Jeffrey Soar, Prabal~Datta Barua, Anthony~R Pisani, and Kathryn Turner.
\newblock Natural language processing in electronic health records in relation to healthcare decision-making: A systematic review.
\newblock \emph{Comput. Biol. Med.}, 155\penalty0 (106649):\penalty0 106649, March 2023.

\bibitem[Iroju et~al.(2015)Iroju, {Department of Computer Science, Adeyemi College of Education, Ondo, Nigeria}, and Olaleke]{Iroju2015-yr}
Olaronke~G Iroju, {Department of Computer Science, Adeyemi College of Education, Ondo, Nigeria}, and Janet~O Olaleke.
\newblock A systematic review of natural language processing in healthcare.
\newblock \emph{Int. J. Inf. Technol. Comput. Sci.}, 7\penalty0 (8):\penalty0 44--50, July 2015.

\bibitem[Jin et~al.(2019)Jin, Dhingra, Cohen, and Lu]{Jin2019-oy}
Qiao Jin, Bhuwan Dhingra, William Cohen, and Xinghua Lu.
\newblock Probing biomedical embeddings from language models.
\newblock In \emph{Proceedings of the 3rd Workshop on Evaluating Vector Space Representations for {NLP}}, pp.\  82--89, Stroudsburg, PA, USA, 2019. Association for Computational Linguistics.

\bibitem[Kanithi et~al.(2024)Kanithi, Christophe, Pimentel, Raha, Saadi, Javed, Maslenkova, Hayat, Rajan, and Khan]{Kanithi2024-ty}
Praveen~K Kanithi, Clément Christophe, Marco A~F Pimentel, Tathagata Raha, Nada Saadi, Hamza Javed, Svetlana Maslenkova, Nasir Hayat, Ronnie Rajan, and Shadab Khan.
\newblock {MEDIC}: Towards a comprehensive framework for evaluating {LLMs} in clinical applications.
\newblock \emph{arXiv [cs.CL]}, September 2024.

\bibitem[Klug et~al.(2024)Klug, Beckh, Antweiler, Chakraborty, Baldini, Laue, Hosch, Nensa, Schuler, and Giesselbach]{Klug2024-gy}
Katrin Klug, Katharina Beckh, Dario Antweiler, Nilesh Chakraborty, Giulia Baldini, Katharina Laue, René Hosch, Felix Nensa, Martin Schuler, and Sven Giesselbach.
\newblock From admission to discharge: a systematic review of clinical natural language processing along the patient journey.
\newblock \emph{BMC Med. Inform. Decis. Mak.}, 24\penalty0 (1):\penalty0 1--13, August 2024.

\bibitem[Kundeti et~al.(2016)Kundeti, Vijayananda, Mujjiga, and Kalyan]{Kundeti2016-fb}
Srinivasa~Rao Kundeti, J~Vijayananda, Srikanth Mujjiga, and M~Kalyan.
\newblock Clinical named entity recognition: Challenges and opportunities.
\newblock In \emph{2016 IEEE International Conference on Big Data (Big Data)}, pp.\  1937--1945. IEEE, December 2016.

\bibitem[Kury et~al.(2020)Kury, Butler, Yuan, Fu, Sun, Liu, Sim, Carini, and Weng]{kury2020chia}
Fabr{'\i}cio Kury, Alex Butler, Chi Yuan, Li-heng Fu, Yingcheng Sun, Hao Liu, Ida Sim, Simona Carini, and Chunhua Weng.
\newblock Chia, a large annotated corpus of clinical trial eligibility criteria.
\newblock \emph{Scientific data}, 7\penalty0 (1):\penalty0 1--11, 2020.

\bibitem[Lee et~al.(2019)Lee, Yoon, Kim, Kim, Kim, So, and Kang]{Lee2019-pb}
Jinhyuk Lee, Wonjin Yoon, Sungdong Kim, Donghyeon Kim, Sunkyu Kim, Chan~Ho So, and Jaewoo Kang.
\newblock {BioBERT}: a pre-trained biomedical language representation model for biomedical text mining.
\newblock \emph{arXiv [cs.CL]}, January 2019.

\bibitem[Li et~al.(2016{\natexlab{a}})Li, Sun, Johnson, Sciaky, Wei, Leaman, Davis, Mattingly, Wiegers, and Lu]{DBLP:journals/biodb/LiSJSWLDMWL16}
Jiao Li, Yueping Sun, Robin~J. Johnson, Daniela Sciaky, Chih{-}Hsuan Wei, Robert Leaman, Allan~Peter Davis, Carolyn~J. Mattingly, Thomas~C. Wiegers, and Zhiyong Lu.
\newblock Biocreative {V} {CDR} task corpus: a resource for chemical disease relation extraction.
\newblock \emph{Database J. Biol. Databases Curation}, 2016, 2016{\natexlab{a}}.
\newblock \doi{10.1093/database/baw068}.
\newblock URL \url{https://doi.org/10.1093/database/baw068}.

\bibitem[Li et~al.(2016{\natexlab{b}})Li, Sun, Johnson, Sciaky, Wei, Leaman, Davis, Mattingly, Wiegers, and Lu]{Li2016-gq}
Jiao Li, Yueping Sun, Robin~J Johnson, Daniela Sciaky, Chih-Hsuan Wei, Robert Leaman, Allan~Peter Davis, Carolyn~J Mattingly, Thomas~C Wiegers, and Zhiyong Lu.
\newblock {BioCreative} {V} {CDR} task corpus: a resource for chemical disease relation extraction.
\newblock \emph{Database (Oxford)}, 2016:\penalty0 baw068, May 2016{\natexlab{b}}.

\bibitem[Luo et~al.(2022)Luo, Lai, Wei, Arighi, and Lu]{DBLP:journals/corr/abs-2204-04263}
Ling Luo, Po{-}Ting Lai, Chih{-}Hsuan Wei, Cecilia~N. Arighi, and Zhiyong Lu.
\newblock Biored: {A} comprehensive biomedical relation extraction dataset.
\newblock \emph{CoRR}, abs/2204.04263, 2022.
\newblock \doi{10.48550/arXiv.2204.04263}.
\newblock URL \url{https://doi.org/10.48550/arXiv.2204.04263}.

\bibitem[Menasalvas et~al.(2016)Menasalvas, Rodriguez-Gonzalez, Costumero, Ambit, and Gonzalo]{Menasalvas2016-zn}
Ernestina Menasalvas, Alejandro Rodriguez-Gonzalez, Roberto Costumero, Hector Ambit, and Consuelo Gonzalo.
\newblock Clinical narrative analytics challenges.
\newblock In \emph{Rough Sets}, Lecture notes in computer science, pp.\  23--32. Springer International Publishing, Cham, 2016.

\bibitem[Niero et~al.(2023)Niero, Souza, Silva, Gumiel, Borges, Piotto, Giavarini, and Oliveira]{Niero2023-ay}
Luiz Henrique~Pereira Niero, João Vitor Andrioli~de Souza, Luciana Martins Gomes~da Silva, Yohan~Bonescki Gumiel, Nícolas~Henrique Borges, Gustavo Henrique~Munhoz Piotto, Gustavo Giavarini, and Lucas Emanuel Silva~e Oliveira.
\newblock Challenges and issues on extracting named entities from oncology clinical notes.
\newblock \emph{J. Health Inform.}, 15\penalty0 (Especial), July 2023.

\bibitem[Névéol et~al.(2018)Névéol, Dalianis, Velupillai, Savova, and Zweigenbaum]{Neveol2018-md}
Aurélie Névéol, Hercules Dalianis, Sumithra Velupillai, Guergana Savova, and Pierre Zweigenbaum.
\newblock Clinical natural language processing in languages other than english: opportunities and challenges.
\newblock \emph{J. Biomed. Semantics}, 9\penalty0 (1):\penalty0 12, March 2018.

\bibitem[{Observational Health Data Sciences} \& {Informatics}(2021){Observational Health Data Sciences} and {Informatics}]{Observational-Health-Data-Sciences2021-ej}
{Observational Health Data Sciences} and {Informatics}.
\newblock {The Book of OHDSI}: Chapter 4 {The Common Data Model}.
\newblock \url{https://ohdsi.github.io/TheBookOfOhdsi/}, January 2021.
\newblock Accessed: 2024-9-2.

\bibitem[Ojha et~al.(2024)Ojha, Doğruöz, Tayyar~Madabushi, Da~San~Martino, Rosenthal, and Rosá]{Ojha2024-gf}
Atul~Kr Ojha, A~Seza Doğruöz, Harish Tayyar~Madabushi, Giovanni Da~San~Martino, Sara Rosenthal, and Aiala Rosá (eds.).
\newblock \emph{Proceedings of the {18th} International Workshop on Semantic Evaluation ({SemEval}-2024)}, Mexico City, Mexico, June 2024. Association for Computational Linguistics.

\bibitem[Peng et~al.(2019)Peng, Yan, and Lu]{Peng2019-tc}
Yifan Peng, Shankai Yan, and Zhiyong Lu.
\newblock Transfer learning in biomedical natural language processing: An evaluation of {BERT} and {ELMo} on ten benchmarking datasets.
\newblock \emph{arXiv [cs.CL]}, June 2019.

\bibitem[Pradhan et~al.(2015)Pradhan, Elhadad, South, Martinez, Christensen, Vogel, Suominen, Chapman, and Savova]{Pradhan2015-bq}
Sameer Pradhan, Noémie Elhadad, Brett~R South, David Martinez, Lee Christensen, Amy Vogel, Hanna Suominen, Wendy~W Chapman, and Guergana Savova.
\newblock Evaluating the state of the art in disorder recognition and normalization of the clinical narrative.
\newblock \emph{J. Am. Med. Inform. Assoc.}, 22\penalty0 (1):\penalty0 143--154, January 2015.

\bibitem[Savova et~al.(2010)Savova, Masanz, Ogren, Zheng, Sohn, Kipper-Schuler, and Chute]{Savova2010-rb}
Guergana~K Savova, James~J Masanz, Philip~V Ogren, Jiaping Zheng, Sunghwan Sohn, Karin~C Kipper-Schuler, and Christopher~G Chute.
\newblock Mayo clinical text analysis and knowledge extraction system ({cTAKES}): architecture, component evaluation and applications.
\newblock \emph{J. Am. Med. Inform. Assoc.}, 17\penalty0 (5):\penalty0 507--513, September 2010.

\bibitem[Shin et~al.(2019)Shin, You, Park, Roh, Kim, Haam, Reich, Blacketer, Son, Oh, and Park]{Shin2019-xw}
Seo~Jeong Shin, Seng~Chan You, Yu~Rang Park, Jin Roh, Jang-Hee Kim, Seokjin Haam, Christian~G Reich, Clair Blacketer, Dae-Soon Son, Seungbin Oh, and Rae~Woong Park.
\newblock Genomic common data model for seamless interoperation of biomedical data in clinical practice: Retrospective study.
\newblock \emph{J. Med. Internet Res.}, 21\penalty0 (3):\penalty0 e13249, March 2019.

\bibitem[Shivade et~al.(2014)Shivade, Raghavan, Fosler-Lussier, Embi, Elhadad, Johnson, and Lai]{Shivade2014-st}
Chaitanya Shivade, Preethi Raghavan, Eric Fosler-Lussier, Peter~J Embi, Noemie Elhadad, Stephen~B Johnson, and Albert~M Lai.
\newblock A review of approaches to identifying patient phenotype cohorts using electronic health records.
\newblock \emph{J. Am. Med. Inform. Assoc.}, 21\penalty0 (2):\penalty0 221--230, March 2014.

\bibitem[Soroush et~al.(2024)Soroush, Glicksberg, Zimlichman, Barash, Freeman, Charney, Nadkarni, and Klang]{Soroush2024-kq}
Ali Soroush, Benjamin~S Glicksberg, Eyal Zimlichman, Yiftach Barash, Robert Freeman, Alexander~W Charney, Girish~N Nadkarni, and Eyal Klang.
\newblock Large language models are poor medical coders — benchmarking of medical code querying.
\newblock \emph{NEJM AI}, 1\penalty0 (5), April 2024.

\bibitem[Stubbs et~al.(2015)Stubbs, Kotfila, and Uzuner]{Stubbs2015-xj}
Amber Stubbs, Christopher Kotfila, and Özlem Uzuner.
\newblock Automated systems for the de-identification of longitudinal clinical narratives: Overview of 2014 {i2b2}/{UTHealth} shared task track 1.
\newblock \emph{J. Biomed. Inform.}, 58 Suppl\penalty0 (Suppl):\penalty0 S11--S19, December 2015.

\bibitem[Sun et~al.(2021)Sun, Yang, Wang, Zhang, Lin, and Wang]{Sun2021-ba}
Cong Sun, Zhihao Yang, Lei Wang, Yin Zhang, Hongfei Lin, and Jian Wang.
\newblock Biomedical named entity recognition using {BERT} in the machine reading comprehension framework.
\newblock \emph{J. Biomed. Inform.}, 118\penalty0 (103799):\penalty0 103799, June 2021.

\bibitem[Wang et~al.(2018)Wang, Singh, Michael, Hill, Levy, and Bowman]{Wang2018-xm}
Alex Wang, Amanpreet Singh, Julian Michael, Felix Hill, Omer Levy, and Samuel Bowman.
\newblock {GLUE}: A multi-task benchmark and analysis platform for natural language understanding.
\newblock In \emph{Proceedings of the 2018 EMNLP Workshop BlackboxNLP: Analyzing and Interpreting Neural Networks for NLP}, pp.\  353--355, Stroudsburg, PA, USA, November 2018. Association for Computational Linguistics.

\bibitem[Wang et~al.(2019)Wang, Pruksachatkun, Nangia, Singh, Michael, Hill, Levy, and Bowman]{Wang2019-vb}
Alex Wang, Yada Pruksachatkun, Nikita Nangia, Amanpreet Singh, Julian Michael, Felix Hill, Omer Levy, and Samuel~R Bowman.
\newblock {SuperGLUE}: A stickier benchmark for general-purpose language understanding systems.
\newblock \emph{Neural Inf Process Syst}, abs/1905.00537:\penalty0 3266--3280, May 2019.

\bibitem[Wu et~al.(2020)Wu, Roberts, Datta, Du, Ji, Si, Soni, Wang, Wei, Xiang, Zhao, and Xu]{Wu2020-gw}
Stephen Wu, Kirk Roberts, Surabhi Datta, Jingcheng Du, Zongcheng Ji, Yuqi Si, Sarvesh Soni, Qiong Wang, Qiang Wei, Yang Xiang, Bo~Zhao, and Hua Xu.
\newblock Deep learning in clinical natural language processing: a methodical review.
\newblock \emph{J. Am. Med. Inform. Assoc.}, 27\penalty0 (3):\penalty0 457--470, March 2020.

\bibitem[Zaratiana et~al.(2023)Zaratiana, Tomeh, Holat, and Charnois]{zaratiana2023gliner}
Urchade Zaratiana, Nadi Tomeh, Pierre Holat, and Thierry Charnois.
\newblock Gliner: Generalist model for named entity recognition using bidirectional transformer, 2023.

\bibitem[Zhang et~al.(2024)Zhang, Zhao, Gao, and Hu]{Zhang2024-yy}
Zhen Zhang, Yuhua Zhao, Hang Gao, and Mengting Hu.
\newblock {LinkNER}: Linking local named entity recognition models to large language models using uncertainty.
\newblock In \emph{Proceedings of the ACM Web Conference 2024}, pp.\  4047--4058, New York, NY, USA, May 2024. ACM.

\end{thebibliography}
\bibliographystyle{iclr2024_conference}

\newpage
\appendix

\section{Appendix}

\subsection{Decoder Model Evaluation}
\label{appendix-decoder-eval}
Evaluating encoder models, such as BERT, for token classification tasks (e.g., NER) is straightforward given that these models process the entire input sequence simultaneously. This allows them to output token-level classifications by leveraging bidirectional context, facilitating a direct comparison of predicted tags against the gold standard labels for each token in the input sequence.

In contrast, decoder-only models, like GPT models, generate responses sequentially, predicting one token at a time based on the preceding context. Evaluating the performance of these models for token classification tasks requires a different approach. First, we prompt the decoder-only LLM with a specific task of tagging the different entity types within a given text. This task is clearly defined to the model, ensuring it understands which types of entities to identify (i.e., conditions, drugs, procedures, etc). An example of the task prompt is shown below.

\begin{Verbatim}[breaklines=true]
## Instruction
Your task is to generate an HTML version of an input text, marking up specific entities related to healthcare. The entities to be identified are: symptom, disorder. Use HTML <span > tags to highlight these entities. Each <span > should have a class attribute indicating the type of the entity. Do NOT provide further examples and just consider the input provided below. Do NOT provide an explanation nor notes about the reasoning. Do NOT reformat nor summarize the input text. Follow the instruction and the format of the example below.
 
## Entity markup guide
Use <span class='symptom' > to denote a symptom.
Use <span class='disorder' > to denote a disorder.
\end{Verbatim}

To ensure deterministic and consistent outputs, the temperature for generation is kept at 0.0. The model then generates a sequential response that includes the tagged entities, as shown in the example below.

\begin{Verbatim}[breaklines=true]
## Input:
He had been diagnosed with osteoarthritis of the knees and had undergone arthroscopy years prior to admission.

## Output:
He had been diagnosed with <span class="disease" >osteoarthritis of the knees</span >and had undergone <span class="procedure" >arthroscopy</span >years prior to admission.
\end{Verbatim}

After the tagged output is generated, it is parsed to extract the tagged entities. The parsed data are then compared against the gold standard labels, and performance metrics are computed as above. This evaluation method ensures a consistent and objective assessment of decoder-only LLM's performance in NER tasks, despite the differences in their architecture compared to encoder models. 

The Universal-NER decoder models series were trained on a specific prompt template the same was used for these to achieve the best performance. 
This is shown in the example below. 
\begin{Verbatim}[breaklines=true]
A virtual assistant answers questions from a user based on the provided text.
USER: Text: {{text}}
ASSISTANT: I've read this text.
USER: What describes {{entity}} in the text?
ASSISTANT: 
\end{Verbatim}

For the GPT4o model, the above html span based prompt template was benchmarked. However to achieve better results, a separate prompt inspired by the universal-ner prompt was used. The scores from this new prompt was used for GPT4o in the benchmark. The prompt used is shown below.

\begin{Verbatim}[breaklines=true]
{%- if is_system_instruction == True -%}
You are a helpful medical LLM that identifies medical entities from the input text.
{%- endif -%}
{%- if is_user_instruction == True -%}
From a given Text, find the entities that describe {{entity}} and return them in a list of strings.
Only output a python list. Do not output anything else like a comment or a suggestion or a note.
For entity spans like 'breast and lung cancer',i.e, entities combined with 'and', output the whole string as a single disease.
Ouptut an empty list if there is no relevant entity.
An example output is: '['entity_text_1', 'entity_text_2']'

Text: {{ text }}
{%- endif -%}
\end{Verbatim}

This was then used to separately query for different entities, which were combined to get the final NER output. Details of the prompting method can be found in our opensource clinical ner benchmark codebase.

\subsection{Common Terminology Label Mapping}
\label{appendix-label-mapping}

The datasets used for the benchmark have numerous entity types. However, the entity labels for the same semantic entities vary across datasets. These entity labels are standardized across datasets using the mapping shown in \ref{tab:Entity Standardization Mapping}.

This mapping was derived by
\begin{itemize}
    \item Referring to the guidelines used while dataset creation
    \item Randomly sampling example entity spans to understand the entity type
\end{itemize}

An important aspect while evaluating models using the mapped entities is that datapoints within datasets like NCBI can also have drug entities which may not have been marked in the ground truth. Therefore, only the existing entity types within a dataset should be used for evaluation. 

\definecolor{EbonyClay}{rgb}{0.121,0.16,0.215}
\begin{table}[t]
\caption{\textbf{Mapping used to standardized dataset entities.}}
\label{tab:Entity Standardization Mapping}
\centering
\begin{tblr}{
  width = \linewidth,
  colspec = {Q[146]Q[208]Q[179]Q[296]Q[106]},
  cell{2}{2} = {fg=EbonyClay},
  cell{2}{3} = {fg=EbonyClay},
  cell{2}{4} = {fg=EbonyClay},
  cell{2}{5} = {fg=EbonyClay},
  cell{3}{3} = {fg=EbonyClay},
  cell{3}{4} = {fg=EbonyClay},
  cell{3}{5} = {fg=EbonyClay},
  cell{4}{3} = {fg=EbonyClay},
  cell{5}{3} = {fg=EbonyClay},
  cell{6}{4} = {fg=EbonyClay},
  cell{7}{4} = {fg=EbonyClay},
  cell{8}{3} = {fg=EbonyClay},
  cell{8}{4} = {fg=EbonyClay},
  hlines,
  vlines,
}

{Standardized\\Label} & NCBI                                                      & CHIA                                                                                                                   & BIORED                     & BC5CDR   \\
Condition             & {CompositeMention,\\DiseaseClass,\\Modifier,\\SpecificDisease} & Condition                                                                                                              & {DiseaseOrPhenotypic-\\Feature} & Disease  \\
Drug                  &                                                           & Drug                                                                                                                   & ChemicalEntity             & Chemical \\
Procedure             &                                                           & Procedure                                                                                                              &                            &          \\
Measurement           &                                                           & Measurement                                                                                                            &                            &          \\
Gene                  &                                                           &                                                                                                                        & GeneOrGeneProduct          &          \\
Gene Variant          &                                                           &                                                                                                                        & SequenceVariant            &          \\
Dropped               &                                                           & {Device, Mood,\\Temporal,\\Negation,\\Observation,\\Qualifier, Scope,\\Reference\_point,\\Person, Value,\\Multiplier, Visit} & {OrganismTaxon,\\CellLine}     &          
\end{tblr}
\end{table}

\begin{figure}[t]
    \centering
        \includegraphics[width=0.6\linewidth]{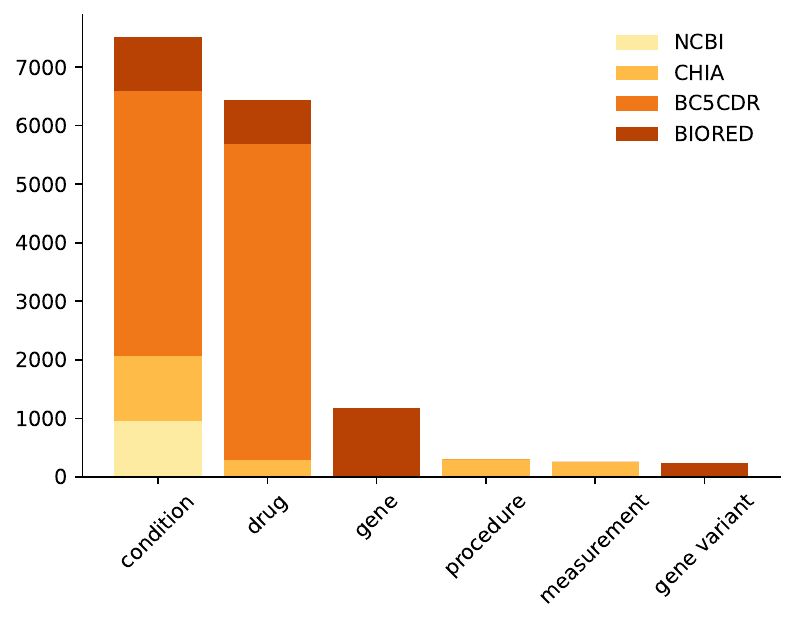}
        \caption{Data Distribution of Clinical Entities}
    \caption{\textbf{Span counts of different entities}. These are the number of entity spans present in the test split of the benchmark datasets}
    \label{fig:span_counts}
\end{figure}

\subsection{Errors of Top Models}
\label{appendix-error-analysis}

\begin{figure}[t]
    \centering
    \begin{subfigure}{0.7\textwidth}
        \includegraphics[width=0.7\linewidth]{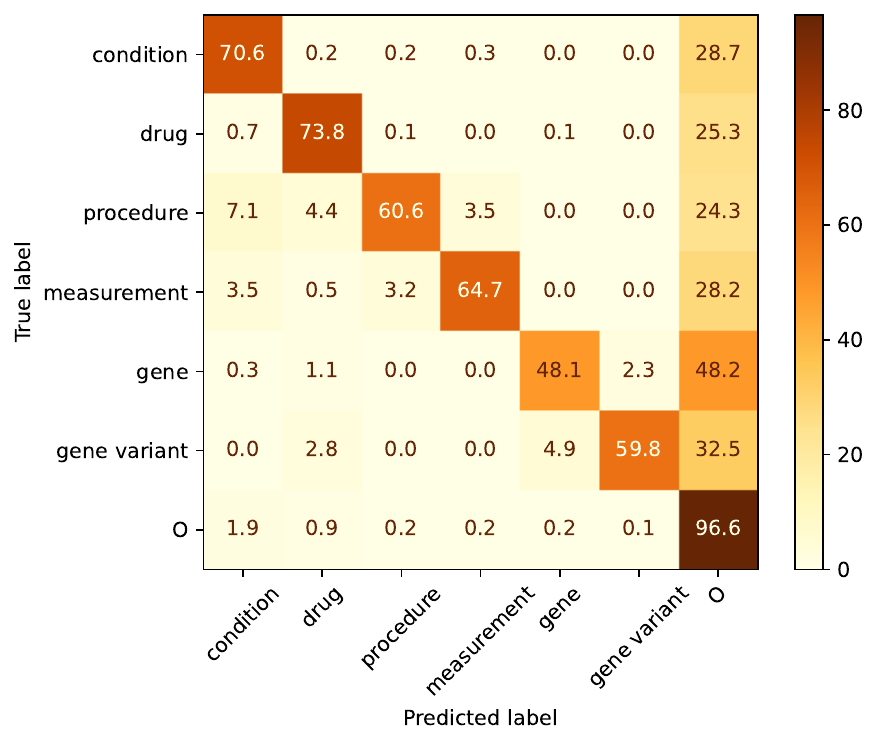}
        \caption{gliner-multitask-large-v0.5}
    \end{subfigure}
    \begin{subfigure}{0.7\textwidth}
        \includegraphics[width=0.7\linewidth]{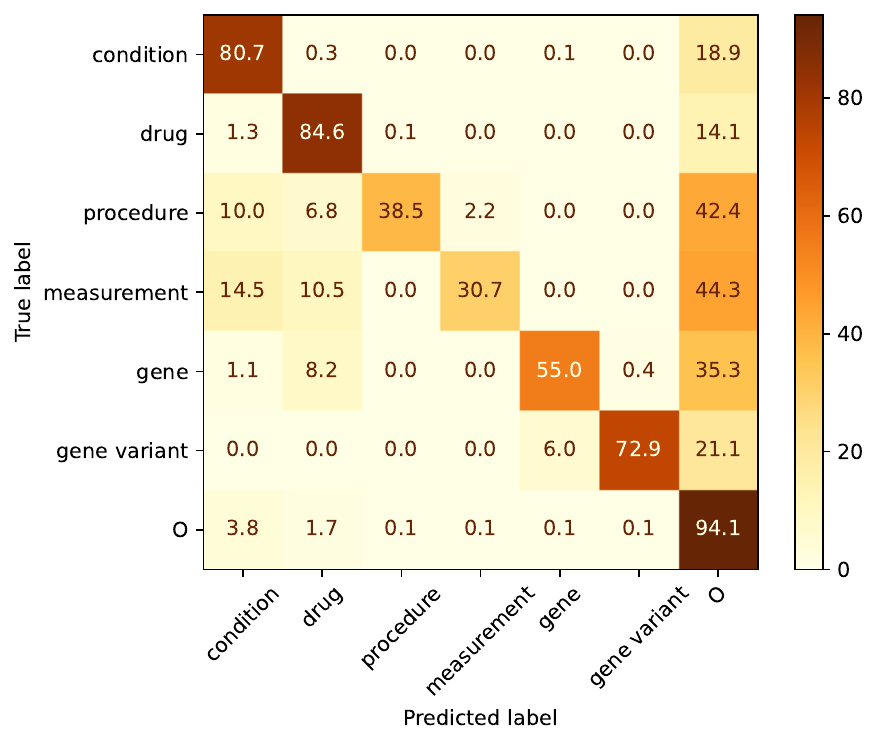}
        \caption{UniNER-7B-type-sup}
    \end{subfigure}
    \begin{subfigure}{0.7\textwidth}
        \includegraphics[width=0.7\linewidth]{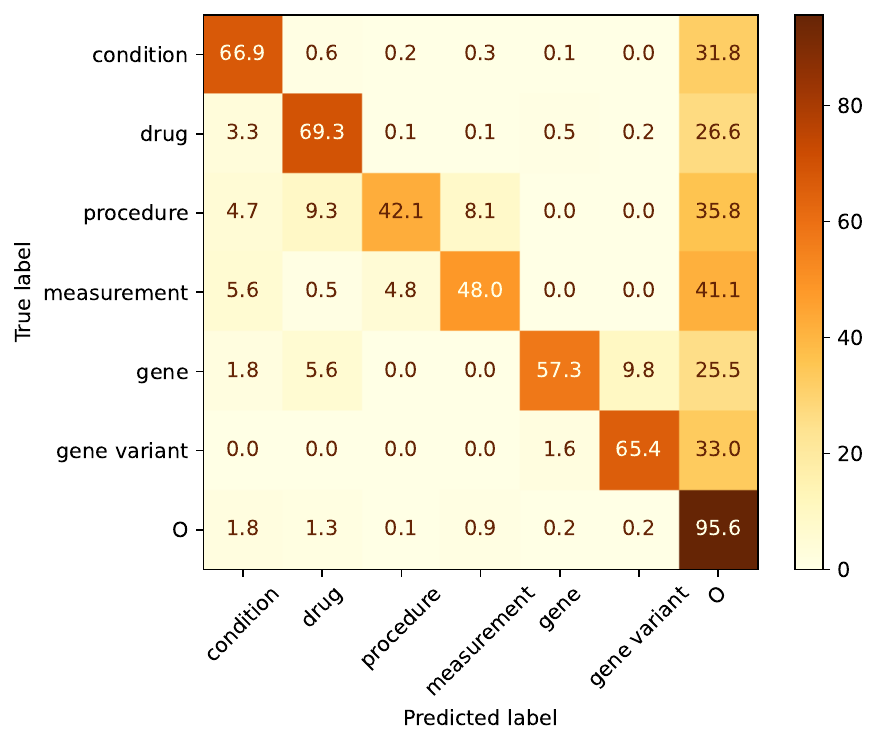}
        \caption{gpt-4o-mini}
    \end{subfigure}
    \caption{\textbf{Confusion Matrices of Top Models}. The numbers represent the percentage of tokens that have been classified/misclassified.}
    \label{fig:confusion-matrices}
\end{figure}

Figure \ref{fig:confusion-matrices} shows the confusion matrices of the top performing models and gpt-4o-mini. The predicted token counts were normalized by the number of token in ground truth(using each model's tokenizer) to obtain the percentage of errors.

\subsection{Detailed Results}
\label{appendix-detailed-results}

\begin{figure}[t]
    \centering
    \includegraphics[width=.6\linewidth]{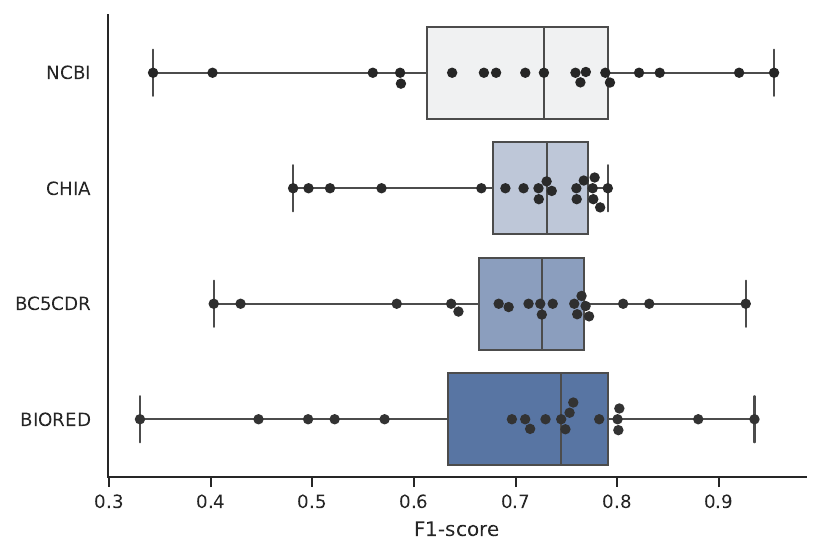}
    \caption{\textbf{Models performance across the various datasets for identifying conditions}. Box plots represent F1-scores of various models across the datasets for condition entities determined using the span-based approach.}
    \label{fig:perf-condition-datasets}
\end{figure}

We present the span and token based results of the leaderboard as of Oct 2024 in table\ref{tab:tabular-results-span} and table\ref{tab:tabular-results-token} respectively. These tables only contain the results on entity types, for dataset resuts, please refer to the leaderboard. 
Figure \ref{fig:perf-condition-datasets} shows the consistency of the entity, Condition, across different datasets. Table \ref{tab:tabular-ranking} shows the effect of metric type on ranking.

\begin{table}[t]
\caption{\textbf{Results on the Leaderboard.} Span metric results of clinical entity types, from the leaderboard as of Oct' 2024.}
\label{tab:tabular-results-span}
\small
\centering
\begin{tabular}{lrrrrrrr}
\toprule
Model & CON. & MEAS. & DRUG & PROC. & GENE & GENE V. & Avg. \\
\midrule
knowledgator/gliner-multitask-large-v0.5 & 77.05 & 66.89 & 76.00 & 58.84 & 62.23 & 52.99 & 65.67 \\
Universal-NER/UniNER-7B-type-sup & 76.92 & 43.69 & 75.13 & 41.30 & 59.88 & 76.72 & 62.27 \\
gliner-community/gliner\_large-v2.5 & 78.25 & 47.60 & 75.74 & 44.67 & 62.08 & 50.20 & 59.76 \\
Universal-NER/UniNER-7B-all & 76.39 & 40.74 & 74.10 & 37.26 & 62.20 & 61.96 & 58.78 \\
urchade/gliner\_large\_bio-v0.1 & 73.83 & 53.62 & 75.01 & 45.30 & 68.38 & 62.74 & 63.15 \\
EmergentMethods/gliner\_large\_news-v2.1 & 74.24 & 26.84 & 75.00 & 51.06 & 62.80 & 62.30 & 58.71 \\
external\_services/gpt-4o-mini-2024-07-18 & 72.50 & 34.12 & 71.00 & 43.94 & 63.64 & 58.85 & 57.34 \\
external\_services/gpt-4o-2024-05-13 & 75.99 & 34.51 & 72.74 & 37.73 & 49.05 & 50.08 & 53.35 \\
urchade/gliner\_large-v2.1 & 71.00 & 44.93 & 73.37 & 50.00 & 58.32 & 63.55 & 60.20 \\
numind/NuNER\_Zero-span & 67.88 & 31.56 & 76.33 & 47.94 & 70.54 & 45.40 & 56.61 \\
Universal-NER/UniNER-7B-type & 68.38 & 43.57 & 69.18 & 38.10 & 55.55 & 39.39 & 52.36 \\
alvaroalon2/biobert\_diseases\_ner & 89.14 & 0.00 & 0.00 & 0.00 & 0.00 & 0.00 & 14.86 \\
meta-llama/Meta-Llama-3-70B-Instruct & 69.17 & 34.13 & 59.05 & 46.98 & 47.65 & 39.42 & 49.40 \\
bioformers/bioformer-8L-ncbi-disease & 86.05 & 0.00 & 0.00 & 0.00 & 0.00 & 0.00 & 14.34 \\
meta-llama/Meta-Llama-3.1-8B-Instruct & 57.09 & 33.38 & 62.63 & 35.91 & 41.65 & 27.10 & 42.96 \\
meta-llama/Meta-Llama-3-8B-Instruct & 59.05 & 26.35 & 57.09 & 28.16 & 49.46 & 27.36 & 41.24 \\
numind/NuNER\_Zero & 46.10 & 28.83 & 61.28 & 31.69 & 59.22 & 33.90 & 43.50 \\
mistralai/Mixtral-8x7B-Instruct-v0.1 & 39.30 & 28.72 & 39.92 & 30.97 & 26.45 & 23.77 & 31.52 \\
\bottomrule
\end{tabular}
\end{table}

\begin{table}[t]
\caption{\textbf{Results on the Leaderboard.} Token metric results of clinical entity types, from the leaderboard as of Oct' 2024.}
\label{tab:tabular-results-token}
\small
\centering
\begin{tabular}{lrrrrrrr}
\toprule
Model & CON. & MEAS. & DRUG & PROC. & GENE & GENE V. & Avg. \\
\midrule
Universal-NER/UniNER-7B-type-sup & 77.43 & 41.85 & 76.81 & 46.36 & 68.00 & 75.59 & 64.34 \\
Universal-NER/UniNER-7B-all & 77.15 & 40.65 & 75.99 & 42.14 & 66.74 & 72.22 & 62.48 \\
knowledgator/gliner-multitask-large-v0.5 & 74.83 & 59.86 & 69.68 & 57.21 & 56.67 & 58.73 & 62.83 \\
gliner-community/gliner\_large-v2.5 & 74.88 & 50.68 & 67.75 & 44.03 & 63.57 & 43.96 & 57.48 \\
urchade/gliner\_large\_bio-v0.1 & 69.99 & 48.08 & 69.93 & 48.28 & 67.42 & 54.23 & 59.66 \\
Universal-NER/UniNER-7B-type & 70.39 & 36.09 & 72.45 & 46.81 & 60.91 & 49.10 & 55.96 \\
external\_services/gpt-4o-2024-05-13 & 73.94 & 20.87 & 66.80 & 38.86 & 58.72 & 57.45 & 52.77 \\
EmergentMethods/gliner\_large\_news-v2.1 & 70.63 & 23.84 & 67.26 & 49.58 & 64.07 & 55.59 & 55.16 \\
urchade/gliner\_large-v2.1 & 66.92 & 38.42 & 66.20 & 48.55 & 55.93 & 56.00 & 55.34 \\
external\_services/gpt-4o-mini-2024-07-18 & 68.68 & 25.55 & 61.47 & 44.46 & 63.15 & 53.34 & 52.78 \\
numind/NuNER\_Zero & 64.21 & 44.15 & 68.73 & 47.10 & 68.37 & 50.90 & 57.24 \\
numind/NuNER\_Zero-span & 62.30 & 36.62 & 66.48 & 47.33 & 71.95 & 46.46 & 55.19 \\
meta-llama/Meta-Llama-3.1-8B-Instruct & 60.05 & 43.15 & 68.99 & 50.49 & 46.91 & 6.28 & 45.98 \\
meta-llama/Meta-Llama-3-8B-Instruct & 63.99 & 32.43 & 65.42 & 40.71 & 52.20 & 9.25 & 44.00 \\
meta-llama/Meta-Llama-3-70B-Instruct & 61.72 & 27.82 & 51.95 & 46.17 & 47.36 & 35.15 & 45.03 \\
alvaroalon2/biobert\_diseases\_ner & 87.87 & 0.00 & 0.00 & 0.00 & 0.00 & 0.00 & 14.65 \\
bioformers/bioformer-8L-ncbi-disease & 81.79 & 0.00 & 0.00 & 0.00 & 0.00 & 0.00 & 13.63 \\
mistralai/Mixtral-8x7B-Instruct-v0.1 & 32.34 & 21.40 & 25.80 & 22.22 & 23.10 & 20.46 & 24.22 \\
\bottomrule
\end{tabular}
\end{table}

\begin{table}[t]
\caption{\textbf{Effect of metrics on Ranking.} The rank is based on average score of clinical entities score. Delta signifies the change in rank on choosing token metric over span metric. }
\label{tab:tabular-ranking}
\small
\centering
\begin{tabular}{lllrrr}
\toprule
Model & Architecture & Type & Span Rank & Token Rank & Delta \\
\midrule
knowledgator/gliner-multitask-large-v0.5 & GLiNER Encoder & zero-shot & 1 & 2 & -1 \\
Universal-NER/UniNER-7B-type-sup & Decoder & fine-tuned & 3 & 1 & 2 \\
gliner-community/gliner\_large-v2.5 & GLiNER Encoder & zero-shot & 5 & 5 & 0 \\
Universal-NER/UniNER-7B-all & Decoder & fine-tuned & 6 & 3 & 3 \\
urchade/gliner\_large\_bio-v0.1 & GLiNER Encoder & zero-shot & 2 & 4 & -2 \\
EmergentMethods/gliner\_large\_news-v2.1 & GLiNER Encoder & zero-shot & 7 & 10 & -3 \\
external\_services/gpt-4o-mini-2024-07-18 & Decoder & zero-shot & 8 & 11 & -3 \\
external\_services/gpt-4o-2024-05-13 & Decoder & zero-shot & 10 & 12 & -2 \\
urchade/gliner\_large-v2.1 & GLiNER Encoder & zero-shot & 4 & 8 & -4 \\
numind/NuNER\_Zero-span & GLiNER Encoder & zero-shot & 9 & 9 & 0 \\
Universal-NER/UniNER-7B-type & Decoder & zero-shot & 11 & 7 & 4 \\
alvaroalon2/biobert\_diseases\_ner & Encoder & fine-tuned & 17 & 17 & 0 \\
meta-llama/Meta-Llama-3-70B-Instruct & Decoder & zero-shot & 12 & 14 & -2 \\
bioformers/bioformer-8L-ncbi-disease & Encoder & fine-tuned & 18 & 18 & 0 \\
meta-llama/Meta-Llama-3.1-8B-Instruct & Decoder & zero-shot & 14 & 13 & 1 \\
meta-llama/Meta-Llama-3-8B-Instruct & Decoder & zero-shot & 15 & 15 & 0 \\
numind/NuNER\_Zero & GLiNER Encoder & zero-shot & 13 & 6 & 7 \\
mistralai/Mixtral-8x7B-Instruct-v0.1 & Decoder & zero-shot & 16 & 16 & 0 \\
\bottomrule
\end{tabular}
\end{table}

\end{document}